\documentclass[sn-mathphys]{sn-jnl}

\jyear{2021}%

\theoremstyle{thmstyleone}%
\theoremstyle{thmstyletwo}%

\theoremstyle{thmstylethree}%

\begin{document}
\title[PCRNet for Ocular Disease Recognition]{Pathology Context Recalibration Network for Ocular Disease Recognition}


\author[1]{\fnm{Zunjie} \sur{Xiao}}

\author[1]{\fnm{Xiaoqing} \sur{Zhang}}

\author*[1,2]{\fnm{Risa} \sur{Higashita}\email{risa@mail.sustech.edu.cn}}
\author*[1,3,4]{\fnm{Jiang} \sur{Liu}\email{liuj@sustech.edu.cn}}

\affil[1]{\orgdiv{Research Institute of Trustworthy Autonomous Systems and Department of Computer Science and Engineering}, \orgname{Southern University of Science and Technology}, \orgaddress{\city{Shenzhen} \postcode{518055},  \country{China}}}

\affil[2]{\orgname{Tomey Corporation}, \orgaddress{\city{Nagoya} \postcode{4510051},  \country{Japan}}}

\affil[3]{\orgdiv{Department of Computer Science and Engineering}, \orgname{Southern University of Science and Technology}, \orgaddress{\city{Shenzhen} \postcode{518055},  \country{China}}}

\affil[4]{\orgdiv{School of Computer Science}, \orgname{University of Nottingham Ningbo China}, \orgaddress{\city{Ningbo} \postcode{315100},  \country{China}}}


\abstract{Pathology context and expert experience play significant roles in clinical ocular disease diagnosis. Although deep neural networks (DNNs) have good ocular disease recognition results, they often ignore exploring the clinical pathology context and expert experience priors to improve ocular disease recognition performance and decision-making interpretability. To this end, we first develop a novel Pathology Recalibration Module (PRM) to leverage the potential of pathology context prior via the combination of the well-designed pixel-wise context compression operator and pathology distribution concentration operator; then this paper applies a novel expert prior Guidance Adapter (EPGA) to further highlight significant pixel-wise representation regions by fully mining the expert experience prior. By incorporating PRM and EPGA into the modern DNN, the PCRNet is constructed for automated ocular disease recognition. Additionally, we introduce an Integrated Loss (IL) to boost the ocular disease recognition performance of PCRNet by considering the effects of sample-wise loss distributions and training label frequencies. The extensive experiments on three ocular disease datasets demonstrate the superiority of PCRNet with IL over state-of-the-art attention-based networks and advanced loss methods. Further visualization analysis explains the inherent behavior of PRM and EPGA that affects the decision-making process of DNNs.}




\maketitle

\section{Introduction}\label{sec1}
With the global aging population, the eye health has become a pronounced public health concern, mainly caused by visual impairment and blindness \cite{swenor2021ageing}. The World Health Organization (WHO) has estimated that approximately 2.2 billion people are suffering from ocular diseases, including cataracts, age-related macular degeneration (AMD), glaucoma, diabetic retinopathy (DR), and myopia~\cite{world2019world,zhang2022machine}. The ophthalmic image-based examination is a commonly used yet effective means for ocular disease screening and diagnosis, which significantly contributes to detecting these ocular diseases early and reducing the ratio of visual impairment and blindness patients.

\begin{figure*}
\centering
       \centerline{\includegraphics[width=0.93\linewidth]{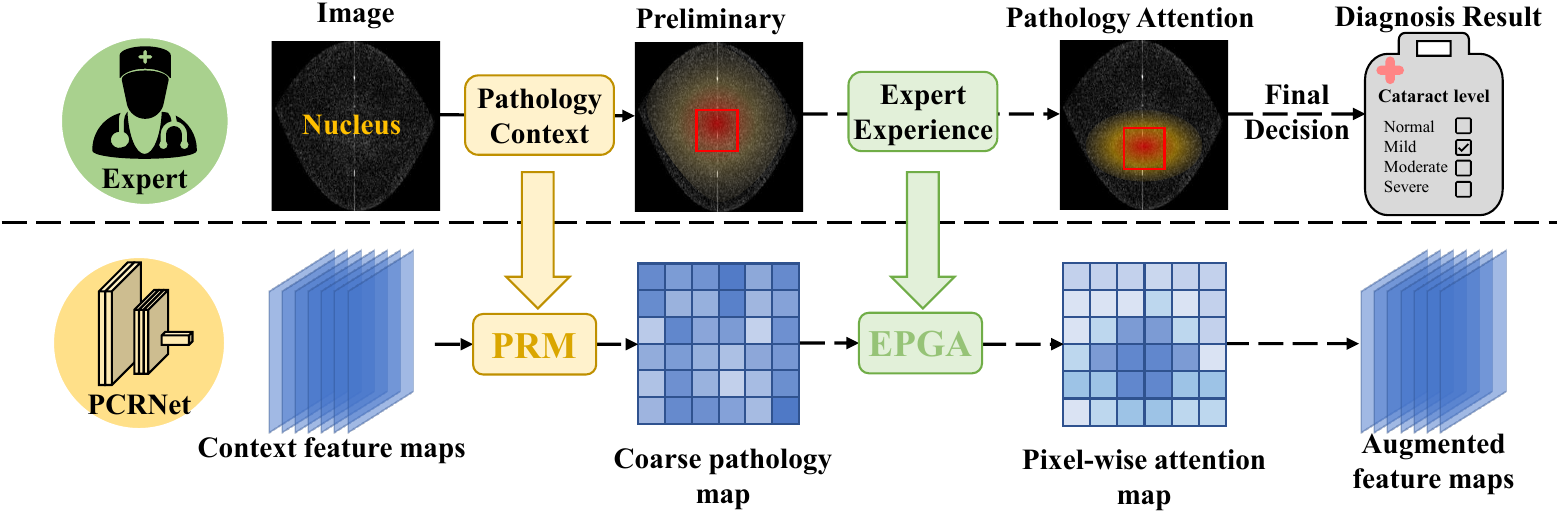}}
   \caption{\textbf{Top}: The flowchart shows how an experienced expert determines the severity level of nuclear cataract (NC) for a subject. First, the expert examines the pathological context of cataracts (e.g., density and location) using the collected AS-OCT images to make a preliminary judgment. Then, drawing on clinical experience, the expert focuses more on the central and lower nuclear regions, which are more closely associated with NC severity. Finally, a conclusive diagnosis is made.
  \textbf{Bottom}: To mimic this diagnostic process, PCRNet first uses the PRM to generate a coarse context map by leveraging pathological knowledge. Then, the EPGA module produces a pixel-wise attention map to further highlight significant pixel context region guided by expert knowledge. Finally, the augmented feature maps are generated by calibrating the context feature maps using the pixel-wise attention map.}
	\label{fig:1}
\end{figure*}

In clinical practice, clinicians typically make diagnosis conclusions for ocular diseases from ophthalmic images, heavily relying on the pathology context and their experience. To be specific, \textit{\romannumeral1)}\textbf{Pathology context.} They are strongly associated with objective pathology changes of ocular diseases, such as location, shape, and texture features, which can be investigated through ophthalmic images and other medical modalities. \textit{\romannumeral2)}\textbf{Expert experience.} It is associated with the extensive clinical training and professional knowledge that clinicians acquired, which also significantly affects the diagnosis conclusions. Moreover, Fig.~\ref{fig:1} (top) illustrates how an experienced expert gives the nuclear cataract (NC) severity level for a subject based on these two clinical priors from AS-OCT images. First, he investigates the pathology changes (e.g., density and location) of cataracts under the collected AS-OCT images and makes a preliminary judgment according to pathology context. Then, according to the expert experience, he pays more attention to the central and down nuclear regions, which are more closely related to NC severity. Finally, he made the final diagnosis conclusions. However, effectively infusing pathology context and expert experience into artificial intelligence–assisted diagnosis techniques for automated ocular disease recognition, particularly in deep neural networks (DNNs), has been rarely explored.

Over the years, attention mechanism has achieved remarkable success across various learning tasks \cite{hu2018squeeze, zhang2024regional, 9195035}, e.g., computer vision and medical image analysis. The critical factor behind the success is that it can enhance the representational capacity of DNNs by guiding them to capture informative context representations. Squeeze-and-excitation (SE) \cite{hu2018squeeze} is a representative channel attention method by building long-dependencies among channels. Efficient channel attention (ECA) \cite{9156697} prompts the idea of SE by modeling the local cross-channel interaction. Self-attention mechanism has recently dominated spatial attention mechanism research direction \cite{zhang2022diffusion}. For example, the non-local (NL) neural network \cite{wang2018non} is proposed to capture long-range dependencies between pixel locations with a self-attention mechanism. Dual attention (DA) \cite{9154612} applies the self-attention method to capture long-range dependencies along spatial and channel dimensions. Multi-scale spatial-
temporal attention Network (MSSTAN) \cite{kong2024multi} integrates spatial attention mechanism in temporal models to enhance the extraction and representation of spatial features in brain regions for functional connectome classification.
Although those attention methods have achieved promising results in medical image analysis, illustrating their design motivation from the perspective of clinical prior (e.g., pathology context and expert experience) infusion remains underexplored.

Moreover, according to Fig.~\ref{fig:1}, we gain two following sights: 1) Only parts of pixel locations in ophthalmic images are associated with the pathology context, similar to informative context representation distributions in feature maps produced by DNNs. 2) Expert experience is employed to further confirm significant pathology context for making the precise diagnosis results, which can be viewed as another form of the pixel-wise context representation bias in feature maps. Motivated by the tight link between these two clinical priors and context representations, a fundamental question naturally arises: \textit{How to infuse these two clinical priors effectively into DNN representations with aid of the attention mechanism design, boosting the ocular disease recognition performance and decision-making interpretability of DNNs?}

To address this problem, this paper introduces a novel Pathology Recalibration Module (PRM) and an Expert Prior Guidance Adapter (EPGA), the first to incorporate pathology context and expert experience priors into DNN representations. Specifically, PRM generates a coarse pathology map through a pixel-wise context compression operator and a pathology distribution concentration operator, by leveraging the merits of pathology context prior. Then, the EPGA fine-tunes the coarse pathology map using the Quantile Statistics Sampling (QSS) method with the aid of expert experience prior exploration. Finally, pixel-wise attention weights recalibrate the spatial positions to highlight/suppress the corresponding pathology context representations. Additionally, we plug PRM and EPGA into modern DNN, constructing Pathology Context Recalibration Network (PCRNet) for automated ocular disease recognition, which provides a substantial performance improvement with minimal computation overhead increase. This paper also offers an intuitive explanation of PCRNet by mining the merits of clinical priors from two perspectives. Furthermore, considering that the collected ophthalmic image distributions of ocular diseases are imbalanced, we develop an improved loss method, termed integrated loss (IL), to boost PCRNet's ocular disease recognition performance by exploiting the potential of sample-wise loss distributions and training label frequency prior.
The main contributions of this paper are summarized as follows:
\begin{itemize}

\item This paper is the first to improve ocular disease recognition performance and decision-making interpretability of DNNs from two perspectives: pathology context and expert experience.

\item This paper develops PRM and EPGA for infusing pathology context and expert experience priors into DNN representations. We construct PCRNet for automated ocular disease recognition by embedding PRM and EPGA into a mainstream DNN. Additionally, we design an integrated loss function (IL) to boost PCRNet’s performance by mining the potential of sample-wise loss distributions and training label frequency prior.

\item The experiments on three datasets demonstrate the superiority of PCRNet and IL over state-of-the-art (SOTA) methods. Visual analysis and ablation study are conducted to interpret the inherent decision-making behavior of PCRNet, proving the effectiveness of our method in allowing DNNs to highlight significant pathology representations and suppress redundant ones, agreeing with the clinical diagnosis mode of clinicians.
\end{itemize}

\section{Related work}
\label{sec:related}

\subsection{Automated Ocular Disease Recognition via Deep Neural Networks} 
Recently, DNNs have achieved excellent performance in ocular disease recognition tasks\cite{li2022annotation,fang2023lcrnet}. Xu et al. \cite{xu2019hybrid} proposed a local-global ensemble CNN framework for cataract diagnosis from fundus images.
Li et al. \cite{9698071} constructed an annotation-free restoration network for improving ocular disease diagnosis results. Gayathri et al. \cite{gayathri2021diabetic} proposed a multipath CNN (M-CNN) to grade DR.
Wu et al.\cite{wu2022expnet} proposed ExpNet for glaucoma classification based on fundus images, and further integrated OCT information through a multi-modality network for improved diagnosis\cite{wu2023gamma}.
Kumar et al. \cite{kumar2021dristi} developed a hybrid deep neural network for automatic DR recognition. Das et al. \cite{das2019multi} proposed a multi-scale deep feature fusion network (MDFF-Net) to classify diabetic macular edema (DME) and age-related macular degeneration (AMD). Li et al.~\cite{li2019canet} proposed a cross-disease attention network (CANet) to distinguish DR. He et al. \cite{9195035} developed a category attention block (CAB) for improving the classification performance of DR by exploring discriminative region-wise context features. Wang et al. \cite{wang2025enhancing} proposed a knowledge-rich vision-language model, RetiZero, to improve performance on rare fundus disease recognition.
Zhang et al. \cite{zhang2022mixed} proposed a mixed pyramid attention network (MPANet) to classify the severity levels of NC. Liu et al.~\cite{liu2025beyond} applied a relational model to detect early dementia based on OCTA images. According to the extensive survey, We found that most existing methods still struggle to achieve SOTA recognition performance by designing complex DNN architectures, inevitably ignoring the intrinsic interpretability requirements from both DNNs and clinical diagnosis. In particular, they have rarely explored the clinical priors of pathology context and expert experience to enhance ocular disease recognition performance and interpretability of DNNs previously.

\subsection{Attention Mechanism} 
Over the years, researchers have widely incorporated attention mechanisms into DNNs, demonstrating their effectiveness in various learning tasks such as image classification and image segmentation~\cite{fu2020scene,guo2022attention,li2019selective,10944285}. Notably, SE \cite{hu2018squeeze} pioneered the learning of channel attention mechanism through global cross-channel interaction. Stye-based recalibration module (SRM) \cite{9008782} incorporated style transfer prior into attention mechanism design using global context representations and channel-wise fully connected layers. Ruan et al. \cite{ruan2020linear} proposed a linear context transform (LCT) block to focus on informative channels. Hu et al. \cite{hu2018gather} developed a gather-excite (GE) block to capture long-range spatial context representation. Recently, Zhang et al.~\cite{zhang2024multiscale} proposed a multiscale channel attention-driven graph dynamic fusion network for mechanical fault diagnosis. Deng et al. \cite{deng2025dense} proposed a time-frequency channel attention (TFCA) mechanism to improve the short-speech speaker recognition results. Self-attention mechanism has emerged as a dominant spatial attention method for guiding DNNs to pay attention where, and various self-attention variants have been developed \cite{guo2022attention,carion2020end}. For example, Ramachandran et al. \cite{ramachandran2019stand} replaced all spatial convolution operations in a CNN with the local self-attention block to enhance performance. Wang et al. \cite{9229188} proposed a parameter-free spatial attention mechanism to capture global spatial relations. Wang et al. \cite{Wang_2017_CVPR} developed a residual attention network incorporating both channel attention and spatial attention to highlight informative global context representations. Convolutional block attention module (CBAM) \cite{woo2018cbam} stacked a channel attention block and a spatial attention block to emphasize/suppress global and local context representations. Park et al. \cite{park2020simple} introduced a bottleneck attention module (BAM) to enable a CNN to focus on salient feature representations through concurrent channel and spatial pathways. Hou et al. \cite{hou2021coordinate} introduced the coordinate attention (CA) mechanism, which embeds spatial information into channel attention. Zhang et al. \cite{zhang2024retfound} proposed a foundation model for fundus disease screening with the help of the self-attention block.

Although existing attention methods have obtained good results through long-range or local dependency modeling across channel and spatial dimensions, their rationality in improving the explanation and performance of DNNs from the clinical diagnosis perspective is still underexplored. Additionally, compared to some previous works that have only incorporated pathology context prior to attention mechanism design by extracting global spatial features or pixel-wise features, this paper attempts to leverage the merits of pathology context and expert experience to improve the ocular disease recognition performance and explanation of DNNs through the external prior infusion block design.

\section{Methodology}
\label{sec:method}

\begin{figure*}
	\centering	
 \centerline{\includegraphics[width=0.98\linewidth]{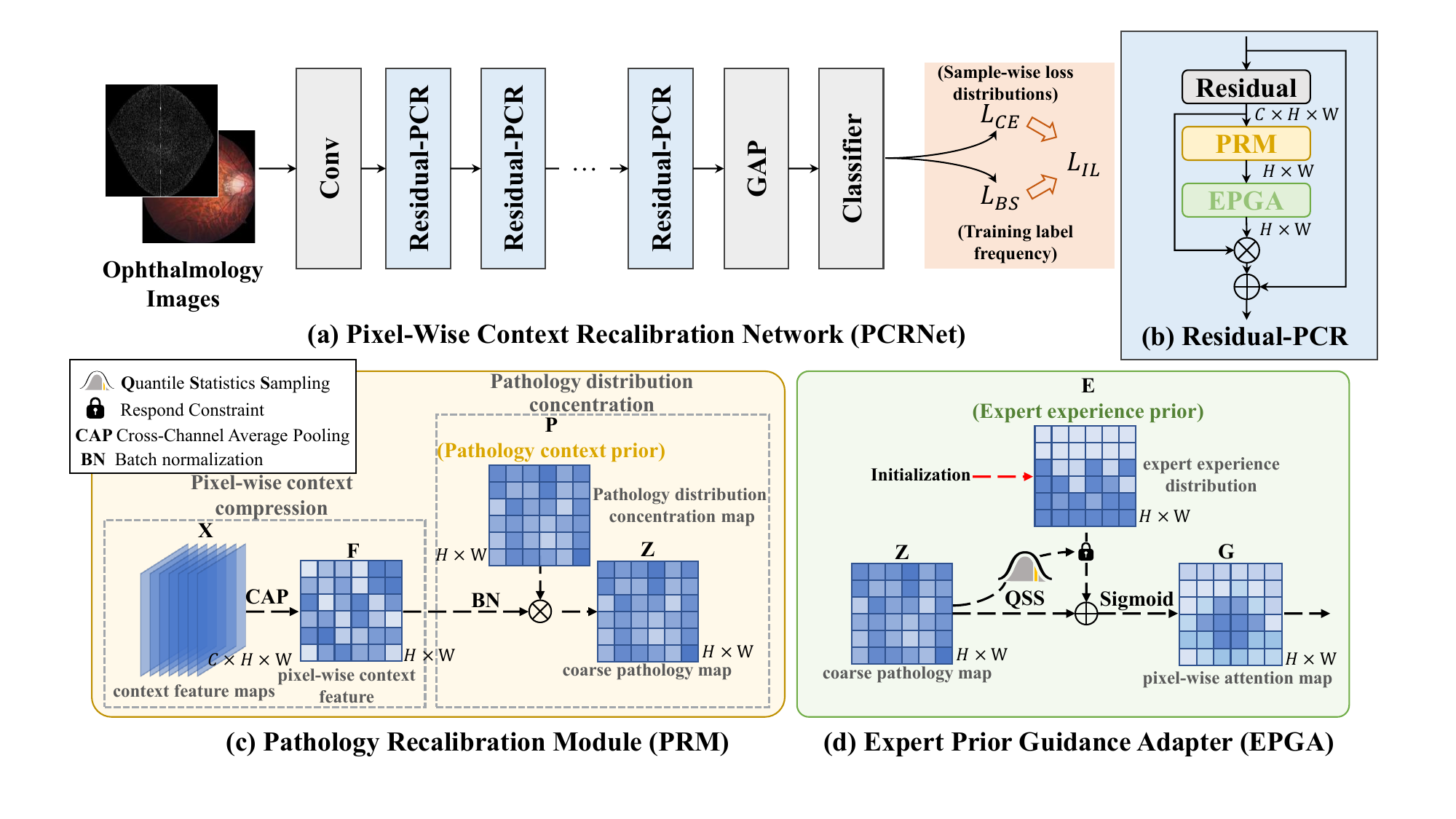}}
	\caption{The framework of the PCRNet with Integrated Loss (IL) is presented to improve ocular disease recognition results based on ophthalmic images. At each stage of PCRNet, we integrate the Pathology Recalibration Module (PRM) and the Expert Prior Guidance Adapter (EPGA) with a residual block to form the Residual-PCR unit. Specifically, the PRM is developed to leverage pathological context prior by combining a pixel-wise context compression operator and pathology distribution concentration operator. Meanwhile, the EPGA further highlights significant pixel-level context regions by effectively incorporating expert experience prior.}
\label{frame}
\end{figure*}

In this section, we first provide an overview of PCRNet. Then, we introduce the specific-designed external prior infusion modules: PRM and EPGA, in detail. Finally, we present the integrated loss (IL), which is applied to improve PCRNet’s ocular disease recognition performance.

\subsection{Network Architecture}
This paper develops a PCRNet for automatic ocular disease recognition, as depicted in Fig.~\ref{frame}(a), which integrates clinical priors from two different perspectives: pathology context and expert experience. Specifically, we apply the Pathology Recalibration Module (PRM) to harness the potential of pathology context prior through a combination of a pixel-wise context compression operator and a pathology distribution concentration operator, as illustrated in Fig.~\ref{frame}(d); an Expert Prior Guidance Adapter (EPGA) is devised to further fine-tunes the coarse pathology map by thoroughly exploiting expert experience prior, as shown in Fig.~\ref{frame}(e).

Fig.~\ref{frame}(a) shows a simple PCRNet framework by taking ResNet as the backbone. It first takes ophthalmic images as inputs, next applies a convolutional layer to extract coarse features, and then processes them through a series of Res-PCR modules (constructed by integrating PRM and EPGA blocks with a residual block as shown in Fig.~\ref{frame}(b)) to produce augmented feature maps. Finally, a global average pooling (GAP) layer generates global feature representations, followed by a softmax classifier to output the ocular disease predictions. Additionally, to improve the ocular disease recognition results of PCRNet, we introduce the integrated loss (IL) by mining the potential of sample-wise loss distributions and training label frequency prior.


\subsection{Pathology Recalibration Module}

PRM generates the coarse pathology map from the context feature maps by mining the pathology context prior. It consists of two main components: pixel-wise context compression operator for aggregating pixel-wise context features along the channel axis, and pathology distribution concentration operator for adaptively capturing pathology context, so as to generate the coarse pathology map.


\textbf{Pixel-wise context compression.}
In order to better extract the pathology context across pixel-wise positions, we use a context descriptor by adopting the common pixel-wise statistics as pixel-wise context features along the channel axis.
Specifically, we select the pixel-wise average as the context descriptor, which can be computed at each pixel location $x(i, j)$ across channels using the cross-channel average pooling (CAP) operator:
\begin{equation}
\mu(i,j) = \frac{1}{C}\sum_{k=1}^{C}x(k,i,j), F= \{\mu(1,1), \mu(1,2),..., \mu(H,W)\},
\label{eq:2}
\end{equation}
where $\mu(i,j)$ denotes pixel-wise average statistic$(i,j)$, $F \in R^{H \times W}$ is pixel-wise context feature. Other pixel-wise features can be obtained through complex cross-channel pooling design; here, we only concentrate on pixel-wise average context features for conceptual clarity and explanation.
In Section~\ref{sec:ablation}, we will verify the superiority of the compression over other operators for gathering context information, such as global average pooling (GAP) for global context feature aggregation in SE.

\textbf{Pathology distribution concentration.} 
Motivated by the observation that pathology context is associated with specific pixel locations under ophthalmic images, we propose an external prior infusion map, pathology distribution concentration map, which differs from existing methods that incorporate pathology context prior into the attention mechanism through global spatial features or pixel-wise features. It employs a learnable tensor to selectively emphasize or suppress features based on their relevance to pathology, which is formulated as follows:
\begin{equation}
    Z = P \odot BN(F), 
\label{eq:4}
\end{equation}
where $P \in R^{H\times W}$ is the pathology distribution concentration map (implicitly denoting objective pathology context prior), $BN$ represents batch normalization (BN), and $\odot$ denotes the pixel-wise production. After re-weighting pixel-wise context features by the Pathology distribution concentration map, the PRM outputs the coarse pathology map $Z \in R^{H\times W}$. 


\subsection{Expert Prior Guidance Adapter.}

The coarse pathology map generated by PRM can only guide DNNs to roughly pay attention to significant pixel-wise context locations, which has limitations in accurately emphasizing pixel-wise context locations. To tackle the shortcomings of PRM, we develop the Expert Prior Guidance Adapter (EPGA), which fully leverages the expert experience prior to further fine-tune the coarse pathology map to highlight more informative pixel-wise context regions. In the EPGA, we initialize the expert experience distribution $E\in R^{H \times W}$ with the Quantile Statistics Sampling (QSS) method to simulate the expert experience in the disease diagnosis condition for obtaining the fine-tuned map. Here, we write our EPGA as follows:
\begin{equation}
    \mu = QSS(Z,\theta),\text{ }G = \sigma(\mu E + Z),
   \label{eq:9}
\end{equation}
where $QSS(Z, \theta)$ represents sampling the $\theta$-th percentile value from $Z$, denoted as $\mu$; $\sigma$ represents the sigmoid function. The sampling value $\mu$ product with $E$ as the bias term to obtain the fine-tuned map $Z^*$, then the gating operator $\sigma$ converts the $Z^*$ to pixel-wise attention map $G$. 

Finally, the intermediate feature maps $X \in R^{C\times H\times W}$ is recalibrated by the pixel-wise attention weights $G$, thus, augmented feature maps $\hat{X} \in R^{C\times H\times W}$ are obtained by:
\begin{equation}
    \hat{X} = X\cdot G.
   \label{eq:10}
\end{equation}

\subsection{Integrated Loss}
Classical cross-entropy (CE) loss is a standard loss for multi-class classification tasks in modern DNNs. Given the mini-batch with training sample number $N$ and class number $K$, it is defined as:  
\begin{equation}
L_{CE} = -\frac{1}{N}\sum_{i=1}^{N}\sum_{j=1}^{K} y_{ij} \log\left(\frac{\exp(z_{ij})}{\sum_{k=1}^{K} \exp(z_{ik})}\right)
\end{equation}
\( y_{ij} \) is the ground truth label for sample \( i \) in class \( j \), and \( z_{ij} \) represents the corresponding predicted logit.  
Although CE loss is effective, its formulation is built on a uniform distribution of all classes with the same sample number, neglecting class imbalance and thereby degenerating the performance of classes with few examples.

To mitigate class imbalance, balanced softmax (BS) loss adjusts predicted logit distributions by mining the potential of training label frequencies, which is formulated as:
\begin{equation}
L_{BS} = -\frac{1}{N}\sum_{i=1}^{N}\sum_{j = 1}^{K} y_{ij} \log\left(\frac{\pi_j \exp(z_{ij})}{\sum_{k}\pi_k \exp(z_{ik})}\right),
\end{equation}
where \(\pi_j\) denotes the training label frequency of class \(j\). It can seen that BS mitigates the issue of gradients dominated by high-frequent classes, promoting low-frequent classes to receive useful gradients. 
However, its reliance solely on training label frequencies and overlooks sample-wise loss characteristics, potentially causing over-biasing. This limitation has been underexplored, which will be validated in subsequent experiments.

Therefore, to explore merits of sample-wise loss distributions and training label frequencies effectively, we propose \textbf{Integrated Loss (IL)} to improve PCRNet’s ocular disease recognition performance, which is written as follows:
\begin{equation}
L_{IL} = (1-\lambda) L_{CE} + \lambda L_{BS},
\end{equation}
where $\lambda$ is a hyper-parameter determined through ablation study to balance the relative contributions of CE and BS. 
The gradient of IL with respect to the logits $z_{ij}$ is given by:
\begin{equation}
\frac{\partial L_{IL}}{\partial z_{ij}} =
\begin{cases}
\frac{1}{N} \sum_{i \in n_k} \tilde{p}_{ik} - 1 & \text{if } j = k \\
\frac{1}{N} \sum_{i \in n_k} \tilde{p}_{ik} & \text{if } j \neq k
\end{cases}
\end{equation}
where $n_k$ is the set of samples in class $k$, and $\tilde{p}_{ik}$ is defined as :
\begin{equation}
\tilde{p}_{ik} = \frac{\exp(z_{ik})}{\sum_{m}\exp(z_{im})} + \lambda \underbrace{\left( \frac{\pi_j \exp(z_{ij})}{\sum_{k}\pi_k \exp(z_{ik})} - \frac{\exp(z_{ij})}{\sum_{k}\exp(z_{ik})} \right)}_{\text{Re-balancing term }\mathcal{R}_{ik}}
\end{equation}
The first part represents the gradient of CE,  while the second part, $\mathcal{R}_{ik}$, serves as a re-balancing term that harmonizes the balance between sample-wise and class-frequent effects through $\lambda$.
To quantify the impact of the re-balancing term, we derive the bounds of \(\mathcal{R}_{ik}\) as follows:
\begin{equation}
\frac{(\pi_j - \pi_{\max}) \exp(z_{ij})}{\pi_{\max} \sum_{k} \exp(z_{ik})} \leq \mathcal{R}_{ik} \leq \frac{(\pi_j - \pi_{\min}) \exp(z_{ij})}{\pi_{\min} \sum_{k} \exp(z_{ik})},
\end{equation}
where \(\pi_{\min}\) and \(\pi_{\max}\) represent the minimum and maximum class frequencies in the training set, respectively. The parameter \(\lambda\) controls the extent of re-balancing, allowing our IL to adjust the influence of CE  and BS contributions by choosing an appropriate \(\lambda\), effectively boosting the ocular disease recognition performance.


\section{Dataset and Experimental Settings}
\label{sec:setting}

\subsection{Datasets}
In this paper, we utilize one private AS-OCT dataset and two publicly available ophthalmic image datasets to comprehensively investigate the effectiveness and generalization ability of our methods. Specifically, the ocular diseases in all three datasets both obey the diagnosis flowchart clinically, as shown in Fig~\ref{fig:1}(a), and all ocular diseases have similar pathological distributions.

\textbf{CASIA2 NC.} It is a clinical AS-OCT dataset of NC with 19,428 AS-OCT images collected through a CASIA2 ophthalmology device. The dataset has four NC severity levels: normal (4891), mild (4842), moderate (5478), and severe (4277). We split the dataset into three disjoint subsets based on the participant level: training (2893/2702/3866/2511), validation(1031/1172/819/749), and testing(967/968/793/1017), confirming that both eyes belonging to a participant fall into the same subset. Table~\ref{dataset_stats_summary} summarizes four severity level distributions of NC in three subsets.

\color{blue}

\begin{table}[ht]
\centering
\caption{Summary statistics across subsets for NC severity levels: normal (4891), mild (4842), moderate (5478), and severe (4277).}
\label{dataset_stats_summary}
\renewcommand{\arraystretch}{1.3}
\begin{tabular}{c|c|c}
\hline
\textbf{Subset} & \textbf{Participants} & \textbf{Images} \\
\hline
Train & 510 / 88 / 138 / 88 & 2,893 / 2,702 / 3,866 / 2,511 \\
Val   & 167 / 46 / 30 / 34  & 1,031 / 1,172 / 819 / 749 \\
Test  & 179 / 34 / 29 / 33  & 967 / 968 / 793 / 1,017 \\
\hline
\end{tabular}
\end{table}
\color{black}

\textbf{LAG.} It is a publicly available fundus image dataset of glaucoma collected by Beijing Tongren Hospital  \cite{ li2019attention}. The dataset has 4854 images: 1,711 glaucoma and 3,143 non-glaucoma. We also split it into three disjoint subsets: training (2,911), validation (971), and testing (972). We also resize the input size of fundus images into $224\times224$.

\textbf{OCTMNIST.} It is a public retinal disease dataset with 109,309 optical coherence tomography (OCT) images (image size: $28\times 28$) for retinal diseases. The dataset contains four retinal disease types: normal, drusen, choroidal neovascularization (CNV), and diabetic macular edema (MDE). We follow the same dataset splitting and image preprocessing methods in \cite{yang2021medmnist}, ensuring a fair experiment comparison. Compared with the other two datasets, the OCTMNIST dataset is a low-resolution image dataset adopted to demonstrate the generalization ability of PCRNet.

\subsection{Evaluation Measures}
We use four commonly-used metrics to verify the overall performance of our method and SOTA attention methods: accuracy (ACC), sensitivity (Sen), F1 score, and kappa coefficient value (Kappa).

\subsection{Strong Baselines}
To demonstrate the superiority of our proposed methods comprehensively, we adopt the following SOTA attention methods, including channel attention, spatial attention, and channel\&spatial attention for comparison: SE, SRM, GE, Non-local \cite{8578911}, coordinate attention (CA) \cite{hou2021coordinate}, CBAM, BAM, and dual attention (DA) \cite{9154612}. 
In addition, we also compared with foundational vision models such as ViT~\cite{dosovitskiy2020image}, Swin-T~\cite{liu2021swin}, ViP~\cite{hou2022vision}, and MLP-Mixer\cite{tolstikhin2021mlp}, as well as specialized cataract classification models including RIRNet~\cite{ZHANG2022102499}, MSSANet~\cite{xiao2024multi}, GCANet~\cite{xiao2021gated}, and CCANet~\cite{zhang2022cca} to further prove its effectiveness.

\subsection{Experimental Settings}
We implement PCRNet and competitive methods with the Pytorch tool. For PCRNet, we adopt two backbone architectures, ResNet18 and ResNet34, which we refer to as PCRNet18 and PCRNet34, respectively.
The stochastic gradient descent (SGD) optimizer is adopted as the optimizer. Training epochs and batch size are set to 150 and 32 accordingly. This paper sets the initial learning rate ($lr$) to 0.0025 and decreases it by a factor of 5 every 20 epochs. We set a fixed $lr$ value to 0.00035 for all models when training epochs over 100. In training, we follow standard data augmentation methods like the random flipping method and the random cropping method for training images. The practical mean channel subtraction is applied to normalize training, validation, and testing datasets. We run all methods on a workstation with an NVIDIA TITAN V (11GB RAM) GPU.

\section{Result Analysis and Discussion}
\label{sec:results}
\subsection{Ablation Study}
In this section, we conduct a series of experiments on the CASIA2 NC dataset with ResNet18 as the backbone to investigate five factors that affect the performance of PCRNet: the initialization of EPGA, the choice of coefficient $\mu$, the implementation of PRM, and the influence of residual-PCR on different stages.

\subsubsection{The impact of Different Initialization Methods for EPGA}
First, we examine the impact of the initialization methods in EPGA in Table~\ref{initial}. Considering diverse expert experience in ocular disease diagnosis, we choose four different preferences as the initialization(left, right, top, bottom), where \texttt{left} denotes setting \( E(i,j) = 1 \) for all \( (i,j) \in \{(i,j) \mid i \leq W/2\} \), and \( E(i,j) = 0 \) otherwise. 
We made a surprising finding that the different initialization methods greatly affect the diagnostic results of the model, corresponding to the differences in pathological distribution. For example, the bottom initialization obtains the best performance in NC grading (the improvement in ACC and Kappa are \textbf{3.14\%} and \textbf{4\%} compared with top initialization, respectively), which is consistent with the pathology distribution observed in \cite{zhang2022adaptive}. Furthermore, the fundus and OCT images also exhibit similar clinical preferences. A more detailed discussion of the initial regions will be presented later in the visualization.

\begin{table} 
\caption{Comparison of different initialization of expert experience distribution $E$ on the CASIA2-NC dataset. (The best are labeled in~\textbf{bold} and second-best are \underline{underlined}.)}
\label{initial}
\centering
\begin{tabular}{c|c|c|c|c}
\hline
Initialization & ACC(\%) & Sen(\%) & F1(\%) & Kappa(\%) \\
\hline
left & 78.91 & \underline{78.05} & \textbf{78.03} & 71.87\\
right & \underline{80.01} & 77.12 & 73.60 & \underline{73.05}\\
top & 78.37 & 77.30 & 77.19& 71.13 \\
bottom & \textbf{81.52} & \textbf{79.25} & \underline{78.02}& \textbf{75.13} \\
\hline
\end{tabular}
\vspace{1mm}
\end{table}

\subsubsection{The Impact of Different Coefficients $\mu$}
In order to maintain contextual information while refining the coarse pathology map, we adjusted the expert experience distribution by coefficients $\mu$ sampled from the coarse pathology map $Z$.
Table~\ref{respond} presents the results obtained for various choices of the coefficient $\mu$. The options include "None", "25th", "50th", and "75th", where "None" indicates that no coefficients $\mu$ are used and the rest indicate the corresponding percentile.
The model appears to not converge without applying the coefficient $ \mu $, which is caused by the original context being overwritten by the expert experience introduced without the constraint.
Additionally, the PCRNet achieved the highest performance with an accuracy of \textbf{81.52\%}, sensitivity of \textbf{79.25\%}, F1 score of \textbf{78.02\%}, and kappa value of \textbf{75.13\%} when set $\mu$ to be 75th percentile.

\begin{table}  

\caption{Comparison of different coefficients $\mu$ with PCRNet18 on CASIA2 NC dataset. (The best are labeled in~\textbf{bold} and second-best are \underline{underlined}.)}
\label{respond}
\centering
\begin{tabular}{c|c|c|c|c}
\hline
$\mu$ & ACC(\%) & Sen(\%) & F1(\%) &Kappa(\%) \\
\hline
None & 50.91& 50.05& 50.03& 50.87\\
25th & 80.45& \underline{79.23}& \textbf{79.18}& 73.87\\
50th & \underline{80.56}& 78.67& \underline{78.08}& \underline{73.91}\\
75th & \textbf{81.52}& \textbf{79.25}&78.02&\textbf{75.13}\\
\hline
\end{tabular}
\vspace{1mm}
\end{table}

\subsubsection{Performance Comparisons of Different Implementations of PCRNet}

Table~\ref{variants} summarizes the experimental results of various PCRNet18 implementations, incorporating different pixel-wise context compression operators and block combinations. The following conclusions can be drawn from the results: 1) MCP and MCP+CAP bring meaningful performance improvement through comparisons to the baseline. However, single CAP (PRM+EPGA) performs better than CMP and CMP+CAP, indicating that pixel-wise context information aggregated by CAP contains different pixel-wise context feature types. 2) Channel (SE) + PRM + EPGA outperforms channel (SE) but is comparable to PRM + EPGA, indicating the proposed method’s effectiveness. . 
3) PRM+EPGA+Conv$3\times3$ slightly performs better than PRM+EPGA, which indicates that the combination of PCRNet and local spatial location modeling is an efficient way to improve performance. However, setting the proper receptive field of the convolution method for spatial attention is challenging. 4) Both PRM and EPGA contribute to practical improvements, and incorporating them enhances overall performance.

\begin{table}  

\caption{Comparison of different variants of PCRNet18 on CASIA2 NC dataset.(The best are labeled in~\textbf{bold} and second-best are \underline{underlined}.)}
\label{variants}
\centering

\begin{tabular}{c|c|c|c|c}
\hline
Method  & ACC(\%) &Sen(\%) & F1(\%) & Kappa(\%) \\
 \hline
ResNet18(baseline) &77.62&75.28&74.51&70.14\\
\hline
CMP &80.77&77.98&75.32&74.05\\
CAP+CMP &80.91&78.06&75.05  &74.21\\ 
channel(SE) &76.56&75.74&76.13&68.79 \\
\hline
PRM+EPGA+Conv$3\times3$  &\underline{81.50}&\underline{80.25}&\underline{79.98}&\textbf{75.25}\\
PRM+EPGA+Conv$5\times5$  &79.39&78.19&78.15&72.45 \\
PRM+EPGA+Conv$7\times7$  &80.19&78.34&77.69&73.41\\ 
channel(SE)+PRM+EPGA  &81.39&\textbf{80.73}&\textbf{80.82}&\underline{75.19}\\
EPGA &77.88&75.56&76.34&70.04\\ 
PRM &80.32&77.60&75.20&73.44\\ 
PRM+EPGA&\textbf{81.52}& 79.25&78.02&75.13\\ 
\hline
\end{tabular}
\begin{flushleft}
\footnotesize *CMP denotes Cross-channel Max Pooling.
\end{flushleft}
\vspace{1mm}
\end{table}
\label{sec:ablation}

\subsubsection{Effects of Residual-PCR on Different Stages of DNNs}
We investigate the impact of replacing the original layers of ResNet18 with Residual-PCR at different stages, one stage at a time. This paper adds Residual-PCR  to the intermediate stages ($Stage\_{1}$, $Stage\_{2}$, $Stage\_{3}$, and $Stage\_{4}$) and presents the classification results in Table~\ref{stage}. We observe that Residual-PCR brings performance improvements when introduced at each architectural stage. Furthermore, the gains from Residual-PCR at different stages are complementary, meaning they can be combined to further enhance network performance, which aligns with the conclusions in SE \cite{Jie2019Squeeze}.

\begin{table}  
\caption{Effect of replacing the original layers of ResNet18 with Residual-PCR at different stages on the CASIA2-NC dataset. (The best are labeled in~\textbf{bold} and second-best are \underline{underlined}.)}
\label{stage}
\centering
\setlength{\tabcolsep}{0.5mm}{
\begin{tabular}{c|c|c|c|c}
\hline
Method & ACC(\%) & F1(\%) & Sen(\%) & Kappa(\%) \\
\hline
ResNet18 & 77.62 & 75.28 & 74.51 & 70.14 \\
+Stage\_1 & 80.24& 78.68& \textbf{78.32} & 73.52\\
+Stage\_2 & 78.64 & 78.14 & 78.32 & 71.57 \\
+Stage\_3 & 80.29 & 78.59 & 78.18 & 73.58 \\
+Stage\_4 & \underline{80.80}& \underline{78.70}& 77.60 & \underline{74.19}\\
+Stage\_all & \textbf{81.52} & \textbf{79.25} & 78.02 & \textbf{75.13} \\
\hline
\end{tabular}}
\vspace{1mm}
\end{table}

\subsubsection{Effects of Gating Operators}

According to Table~\ref{gate}, we observe that exchanging the sigmoid for tanh and ReLU dramatically worsens the NC classification performance, while using softmax is slightly worse than sigmoid. The following reasons might account for the results: 1) the attention weights generated by tanh between -1 and 1, while negative attention weights negatively affect a network's parameter updating. 2) ReLU sets part of attention weights to 0, resulting in parts of spatial information being neglected. 3) softmax captures the global pixel-wise context information between spatial locations, which introduces spatial information redundancy, presented as that the gap between the most attention weights is small. This also suggests that it is vital to construct the gating operator in EPGA carefully.

\begin{table}  
\caption{Effect of gating operators for EPGA with PCRNet18 on CASIA2 NC dataset. (The best are labeled in~\textbf{bold} and second-best are \underline{underlined}.)}
\label{gate}
\centering
\scalebox{1.0}{
\begin{tabular}{c|c|c|c|c}
\hline
Method(\%)  & ACC(\%) & Sen(\%)& F1(\%) & Kappa(\%) \\
 \hline
ReLU &79.67 &77.22 & 75.99&72.38   \\
Tanh &78.85 &77.72 &77.70&71.75   \\
Softmax &\underline{80.48}&\underline{78.60}& \underline{77.88}&\underline{73.79}\\
Sigmoid&\textbf{81.52}& \textbf{79.25}&\textbf{78.02}&\textbf{75.13} \\
 \hline
\end{tabular}}
\vspace{1mm}
\end{table}

\subsection{Comparisons with SOTA Attention Methods}

\paragraph{Results on CASIA2 NC Dataset.}
We first compare our method with other SOTA attention methods on the CASIA2 NC dataset by using ResNets (ResNet18 and ResNet34) as backbones. Table~\ref{tab:2}(Left) summarizes the NC classification results. It can be observed that our PCRNet continuously improves the classification performance over other attention methods under similar or less model complexity. 
Specifically, although DA and Non-local are \textbf{6.87\%} larger in parameters and \textbf{6.38\%} larger in computation, PCRNet outperforms DA and Non-local by over \textbf{4.3\%} on four evaluation measures. For instance, PCRNet18 obtains \textbf{4.46\%} gains of accuracy compared with DA under ResNet18. The possible reason for explaining the results is that DA and Non-local belong to the self-attention method, which introduces redundant information by capturing long-range dependencies between spatial locations that has been discussed by previous works. With comparable or slightly more complexity than GE, PCRNet achieves over \textbf{4.5\%} and \textbf{5.9\%} gains of accuracy and kappa accordingly. 
Overall, the results demonstrate the effectiveness of PCRNet through comparisons to other competitive attention methods.

\begin{table*}
\caption{Performance and complexity comparisons of our PCRNet and SOTA attention methods on three ocular disease datasets (CASIA2 NC, LAG, OCTMNIST) in terms of ACC(\%), Sen(\%), F1(\%), Kappa(\%), Params(M) and GFLOPs. (The best are labeled in~\textbf{bold} and second-best are \underline{underlined}.)}

\label{tab:2}
\centering
\scalebox{0.50}{
\begin{tabular}{c|c|c|c|c|c|c|c|c|c|c|c|c|c|c}
\hline
\multirow{2}{*}{Backbone}  & \multicolumn{4}{c|}{CASIA2 NC}& \multicolumn{4}{c|}{LAG}  & \multicolumn{4}{c|}{OCTMNIST}  & \multirow{2}{*}{Params}&\multirow{2}{*}{GFLOPs}\\
 \cline{2-13}
 & ACC &Sen&F1&Kappa& ACC &Sen&F1&Kappa& ACC &Sen&F1&Kappa&&\\
 \hline
 ResNet18 \cite{he2016deep}&77.62&75.28&74.51&70.14&93.83&93.51&93.28&86.56&76.20&76.20&73.61&68.27&11.171&0.458\\
+SE \cite{Jie2019Squeeze} &76.56&75.74&76.13&68.79&93.93&93.39&93.36&86.71&77.50&77.50&75.14&70.00&11.260&0.458\\
+SRM \cite{9008782}&78.16&76.66&76.30&70.78&94.03&93.80&93.51&87.02&75.30 &75.30&72.36&67.07&11.175&0.458\\
+GE \cite{hu2018gather}&76.80&76.29&76.56&69.14&93.42&92.46&92.74&85.49&\underline{79.30}&\underline{79.30}&\underline{77.23}&\underline{72.40}&11.454&0.458\\
+Non-local \cite{8578911}& 73.16& 70.98&69.71&63.89&89.20&88.14&88.17&76.33&74.70&74.70&70.50&66.27& 11.956&0.489\\
+CA \cite{hou2021coordinate}&77.06&74.96&73.90&69.21&93.21&92.90&92.61&85.23&78.10&78.10&74.49&70.80&11.307&0.459\\
+CBAM \cite{woo2018cbam} &\underline{79.20}&\underline{77.62}&\underline{77.20}&\underline{72.14}&\underline{94.86}&\underline{94.50}&\underline{94.38}&\underline{88.77}&78.20&78.20&75.22&70.93&11.260&0.458\\
+BAM \cite{park2020simple}&77.89&76.95&77.02&70.50&93.42&93.12&92.84&85.68&77.00&77.00&73.26&69.33&11.194&0.459\\
+DA \cite{9154612}&77.06&74.95&73.90&69.21&92.80&91.65&92.04&84.08&76.70&76.70&74.02&68.93& 11.956&0.489\\
\hline
PCRNet18& \textbf{81.52}& \textbf{79.25}& \textbf{78.02}& \textbf{75.13}& \textbf{95.58}& \textbf{95.12}& \textbf{95.15}& \textbf{90.31}& \textbf{82.40}& \textbf{82.40}& \textbf{80.85}& \textbf{76.53}& 11.175& 0.458\\
\hline
\hline

ResNet34 \cite{he2016deep} &76.64&75.66&75.56&68.84&93.62&92.75&92.98&85.96&77.30&77.30&74.24&69.73&21.278&0.938\\
+SE \cite{Jie2019Squeeze} &79.49&\underline{76.59}&73.62&72.29&93.72&93.30&93.15&86.30&77.30&77.30&74.28&69.73&21.440&0.939\\
+SRM \cite{9008782}&\underline{79.76}&76.21&73.04&\underline{72.63}&93.62&\underline{93.55}&93.09&86.18&76.20&76.20&72.78&68.27&21.287&0.938\\
+GE \cite{hu2018gather}&76.61&75.73&\underline{75.91}&68.82&\underline{93.83}&93.24&\underline{93.24}&\underline{86.48}&\underline{78.90}&\underline{78.90}&\underline{76.93}&\underline{71.87}&21.794&0.939\\
+Non-local \cite{8578911}&73.56&71.48&70.2&61.06&86.21&84.97&84.73&69.85&68.30&68.30&61.77&57.33&22.698&1.000\\
+CA \cite{hou2021coordinate}&74.66&73.53&72.80&66.19&92.70&91.77&91.97&83.93&77.30&77.30&73.98&69.73&21.525&0.941\\
+CBAM \cite{woo2018cbam} &73.65&72.53&72.04&64.86&93.62&92.62&92.96&85.92&78.00&78.00&75.27&70.67&21.440&0.939\\
+BAM \cite{park2020simple}&74.15&72.86&72.47&65.48&93.42&92.39&92.73&85.47&77.20&77.20&74.90&69.60&21.302&0.940\\
+DA \cite{9154612}&75.14&72.96&71.43&66.60&90.64&89.32&89.66&79.32&78.80&78.80&76.33&71.73&22.698&1.000\\
\hline
PCRNet34& \textbf{81.20}& \textbf{79.74}& \textbf{79.54}& \textbf{74.82}& \textbf{94.86}& \textbf{94.70}& \textbf{94.41}& \textbf{88.81}& \textbf{81.70}& \textbf{81.70}& \textbf{79.74}& \textbf{75.60}& 21.286& 0.938\\
\hline
\end{tabular}}
\end{table*}

\paragraph{Results on LAG Dataset.}
Table~\ref{tab:2}(Middle) offers the glaucoma detection results of our PCRNet and SOTA attention methods. We see that PCRNet consistently improves the glaucoma detection performance over SOTA attention methods. Remarkably, our PCRNet18 outperforms Non-local by above absolute \textbf{6.0\%} on accuracy, while Non-local is \textbf{6.87\%} larger in parameters and \textbf{6.38\%} larger in computational cost. With less complexity than CBAM and BAM, PCRNet achieves over \textbf{1.6\%} gain of accuracy and \textbf{3.9\%} gain of kappa. Compared with SRM under similar parameters and computational cost, PCRNet also achieves over \textbf{1.0\%} gain on four evaluation measures. 


\paragraph{Results on OCTMNIST Dataset.}
Table~\ref{tab:2}(Right) presents the retinal disease classification results of PCRNet and SOTA attention methods on the OCTMNIST dataset. Different from the image size 224$\times$224 of the other two ophthalmic image datasets, the image size of the OCTMNIST dataset is 28 $\times$ 28, which is used to demonstrate that PCRNet can focus on significant feature map regions effectively to improve the performance on low-resolution medical image dataset with fewer parameters. We see that using fewer parameters and computations than other advanced attention methods, PCRNet18 obtains absolute over \textbf{3.1\%} on four evaluation measures on ResNet18. Remarkably, compared with CA and SRM, PCRNet18 significantly outperforms them by over \textbf{4.3\%} of accuracy, \textbf{6.36\%} of F1, and \textbf{5.73\%} of kappa, respectively.

The results in Table~\ref{tab:2} demonstrate the generalization ability and superiority of our PCRNet, keeping consistent with our expectations.

\subsection{Comparisons with Advanced Loss Methods}

Table~\ref{loss} lists the comparison results of four competitive classification losses: focal loss (FL) \cite{lin2017focal},  label-distribution-aware margin loss (LADM) \cite{cao2019learning}, cross-entropy loss (CE), balanced softmax loss (BS) \cite{ren2020balanced}, and our integrated loss (IL) with different hyper-parameter $\lambda$ settings. 
We can observe that the IL$(\lambda=0.5)$ achieves the best performance, with \textbf{2.73\%} gain of F1 score compared with  CE and BS, Focal, and LADM, which proves the complementary between CE and BS. 
Besides, IL is the combination of CE and BS, and its performance is sensitive to the choice of hyper-parameter $\lambda$, with the best gain when $\lambda=0.5$. Therefore, the $\lambda$ is fixed to 0.5 in the rest experiments.

\begin{table}  
\caption{Performance comparison of IL and Advanced Loss Methods for PCRNet18 on the CASIA2 NC dataset. (The best are labeled in~\textbf{bold} and second-best are \underline{underlined}.)}
\label{loss}
\centering
\begin{tabular}{c|c|c|c|c}
\hline
Method & ACC(\%) & Sen(\%)& F1(\%) & Kappa(\%) \\
 \hline
CE & \underline{81.52}& 79.25& 78.02& \underline{75.13}\\
BS\cite{ren2020balanced} &77.49&76.99&77.19&70.06 \\
Focal & 80.56& 78.79& 78.18&73.91 \\
LADM\cite{cao2019learning} & 78.51& 77.53& 77.38& 71.32\\
\hline
IL($\lambda$=0.2)&80.69& \underline{79.27}& \underline{79.18}& 74.16\\
IL($\lambda$=0.5)&\textbf{81.87}&\textbf{80.76}&\textbf{80.75}&\textbf{75.75} \\
IL($\lambda$=0.8)&79.47&78.86&78.93&72.65\\
 \hline
\end{tabular}

\end{table}

\subsection{Comparisons with SOTA Deep Neural Networks}

\begin{table}[t]
\caption{Performance comparisons of PCRNet18 and SOTA Deep
Neural Networks on CASIA2 NC dataset in terms of ACC(\%), Sen(\%), F1(\%), Kappa(\%). (The best are labeled in~\textbf{bold} and second-best are \underline{underlined}.)}
\
\label{NC}
\centering

\begin{tabular}{c|c|c|c|c}
\hline
Method & ACC &Sen&F1&Kappa\\
 \hline
ViT\cite{dosovitskiy2020image}& 69.64& 67.81& 67.23& 59.23\\
Swin-T~\cite{liu2021swin}& 74.34& 72.33& 71.01&65.55 \\
ResMLP\cite{touvron2021resmlp}& 74.95& 72.99& 72.04& 66.34\\
ViP\cite{hou2022vision}& 72.52& 70.75& 70.28& 63.18\\
MLP-Mixer& 73.83& 72.23& 71.71& 64.93\\
RIRNet~\cite{ZHANG2022102499}& 76.58& 75.60& 75.37& 68.73\\
MSSANet~\cite{xiao2024multi}& 78.40& 76.89& 76.56& 71.08\\
CCANet~\cite{zhang2022cca}& 79.33& 78.51& 78.64& 72.44\\
GCANet~\cite{xiao2021gated}& 79.76& 77.75& 76.97& 72.82\\
\hline
PCRNet18(CE)&\underline{81.52}&\underline{79.25}&\underline{78.02}&\underline{75.13}\\
PCRNet18(IL)&\textbf{81.87}&\textbf{80.76}&\textbf{80.75}&\textbf{75.75}\\
\hline
\end{tabular}
\end{table}

\begin{table}[t]
\caption{Comparative analysis of prediction differences between PCRNet18 and state-of-the-art deep neural networks on the CASIA2 NC dataset. (p-value $\le$ 0.05 indicates a statistically significant improvements.)}
\
\label{t-test}
\centering

\begin{tabular}{c|c}
\hline
Method & p-value\\
 \hline
PCRNet18 VS. ViT& $\leq$1e-5\\
PCRNet18 VS. Swin-T& $\leq$1e-5\\
PCRNet18 VS. ResMLP& $\leq$1e-5\\
PCRNet18 VS. ViP& $\leq$1e-5\\
PCRNet18 VS. MLP-Mixer& $\leq$1e-5\\
PCRNet18 VS. RIRNet& 0.0431\\
PCRNet18 VS. MSSA& 0.0452\\
PCRNet18 VS. CCANet& $\leq$1e-5\\
PCRNet18 VS. GCANet& $\leq$1e-5\\
\hline
\end{tabular}
\end{table}

\begin{table}[t]
\caption{Performance comparisons of PCRNet18 and SOTA Deep
Neural Networks on LAG dataset in terms of ACC(\%), Sen(\%), F1(\%), Kappa(\%). (The best are labeled in~\textbf{bold} and second-best are \underline{underlined}.)}
\
\label{LAG}
\centering
\begin{tabular}{c|c|c|c|c}
\hline
Method  & ACC &Sen&F1&Kappa\\
 \hline
 MDFF \cite{DAS2019101605} &93.90&91.80&91.50&83.0\\
 CNN \cite{li2019large}& 89.20 & 90.60&-- &--\\
 EfficientNet~\cite{tan2019efficientnet} &94.14&93.95&93.63 &87.25\\
      RIRNet~\cite{ZHANG2022102499}&93.11&92.75&92.50&85.00\\
    DR-Net \cite{guo2020dense}&93.42&92.66&92.77&85.54 \\ 
    MSCA-Net~\cite{fu2023rmca}&93.00&92.01&92.29&84.58 \\
     FIT-Net~\cite{chen2023fit}&92.28&91.39&91.52&83.04\\
 ViT \cite{dosovitskiy2020image} &89.92&89.56&89.09&78.18\\
     Swin-T ~\cite{liu2021swin}&91.80&90.95&91.17&82.34\\
     ResMLP\cite{touvron2021resmlp} &91.98&90.62&91.11&82.22\\
\hline
PCRNet18(CE)&\textbf{95.58}&\underline{95.12}&\underline{95.15}&\underline{90.31}\\
PCRNet18(IL)&\underline{95.57}&\textbf{95.65}&\textbf{95.19}&\textbf{90.39}\\
\hline
\end{tabular}
\end{table}

\begin{table}[t]
\caption{Performance comparisons of PCRNet18 and SOTA Deep
Neural Networks on OCTMNIST dataset in terms of ACC(\%), Sen(\%), F1(\%), Kappa(\%). (The best are labeled in~\textbf{bold} and second-best are \underline{underlined}.)}
\
\label{OCTMNIST}
\centering
\begin{tabular}{c|c|c|c|c}
\hline
Method & ACC &Sen&F1&Kappa\\
 \hline
 ViT\cite{dosovitskiy2020image}& 71.40& 71.40& 66.16& 61.87\\
ConvNeXt \cite{liu2022convnet}& 72.60& 72.60& 67.83& 63.47\\
MedViT~\cite{manzari2023medvit}& 78.20&- & -& -\\
Google AutoML \cite{bisong2019google}& 77.10& -& -& -\\ 
RIRNet~\cite{ZHANG2022102499}& 81.10& 81.10& 79.49& 74.80\\
PVTV2~\cite{wang2022pvt}& 79.50& 79.50& 77.77& 72.67\\
PSANet~\cite{9229188}& 76.10& 76.10& 72.60&-\\
\hline
PCRNet18(CE)&\underline{81.70}&\underline{81.70}&\underline{79.74}&\underline{75.60}\\
PCRNet18(IL)&\textbf{83.00}&\textbf{83.00}&\textbf{81.67}&\textbf{77.33}\\
\hline
\end{tabular}
\end{table}

\begin{table}[t]
\caption{The out-of-distribution (OOD) evaluation of PRCNet18 and SOTA DNNs on an extra NC severity classification dataset (All the models pre-trained on the CASIA2 NC dataset without fine-tuning).}
\
\label{ood}
\centering

\begin{tabular}{c|c|c|c|c}
\hline
Method & ACC &Sen&F1&Kappa
\\
 \hline
ViT\cite{dosovitskiy2020image}&    41.96&26.06&23.04&7.24\\
Swin-T~\cite{liu2021swin}&    38.69&30.45&31.13&20.44\\
ResMLP\cite{touvron2021resmlp}&    37.65&24.97&27.99&17.75\\
ViP\cite{hou2022vision}&    \textbf{47.47}&\textbf{34.53}&\underline{36.45}&\underline{27.42}\\
MLP-Mixer&    43.16&29.99&36.46&27.42\\
RIRNet~\cite{ZHANG2022102499}&    24.44&15.43&17.72&8.35\\
MSSANet~\cite{xiao2024multi}&    31.99&22.00&28.06&17.93\\
CCANet~\cite{zhang2022cca}&    35.42&24.87&30.52&16.09\\
GCANet~\cite{xiao2021gated}&    24.55&16.24&16.66&11.40\\
\hline
 PCRNet18& \underline{47.17}& \underline{34.15}& \textbf{40.45}&\textbf{27.66}\\
 \hline
\end{tabular}
\end{table}

\paragraph{Results on CASIA2 NC Dataset.}
Table~\ref{NC} offers the classification results of our PCRNet and recent advanced DNNs on CASIA2 NC dataset. We observe that our PCRNet generally achieves the best performance among all methods. For example, our PPCA outperforms transformer and MLP-like architectures like ViT, Swin-T, and MLP-Mixer by 6.57\% in accuracy and 5.98\% in F1, respectively. Especially, compared with previous cataract classification methods, RIRNet, MSSANet, CCANet and GCANet, PCRNet also achieves over 2.31\% kappa improvement. 
Besides, integrated loss (IL) also further boosts NC recognition performance of PCRNet. 
Moreover, Table~\ref{t-test} presents a comparative analysis of prediction differences between PCRNet and state-of-the-art deep neural networks on the CASIA2 NC dataset. The statistical analysis results show that PCRNet achieves statistically significant improvements compared to SOTA methods.

\paragraph{Results on LAG and OCTMNIST.}
Table~\ref{LAG} and Table~\ref{OCTMNIST} list the results of PCRNet and other competitive DNNs on LAG and OCTMNIST dataset, respectively. Due to the input image size problem, some SOTA DNNs are not applicable to OCTMNIST datasets. We can see that our PCRNet generally performs better than comparable DNNs, and IL further boosts its performance, keeping consistent with our motivations. For example, PCRNet with hybrid loss outperforms Swin-T, ViT, and RIRNet by absolute over \textbf{2.47\%} of accuracy and F1 on the LAG dataset. 
In Table~\ref{OCTMNIST}, PCRNet with IL  also obtains over \textbf{11\%} increase in terms of ACC, Sen, F1, and Kappa compared with ViT. 
All in all, the results in Table~\ref{NC} - \ref{OCTMNIST} demonstrate the superiority and generalization of our PCRNet and IL.

\paragraph{Out-of-distribution Test in NC Severity Classification Task.}
To further demonstrate the real-world applicability of our PCRNet, we introduce an out-of-distribution (OOD) dataset of AS-OCT images collected with varying NC severity, noting that this dataset does not include normal cases. The results in Table~\ref{ood} show that PCRNet achieves stable performance compared to others. Specifically, PCRNet outperforms Swin-T, ResMLP, and MSSANet, with an increase of over 9\% in F1 score, highlighting its robustness and effectiveness.

\subsection{Visualization and Interpretability}
This section visualizes the pathology distribution concentration map $P$, expert experience distribution $E$, and pixel-wise attention map $G$ of PCRNet, aiming to enhance the interpretability of deep networks.

\begin{figure}	 
	\centering	
 \centerline{\includegraphics[width=0.8\linewidth]{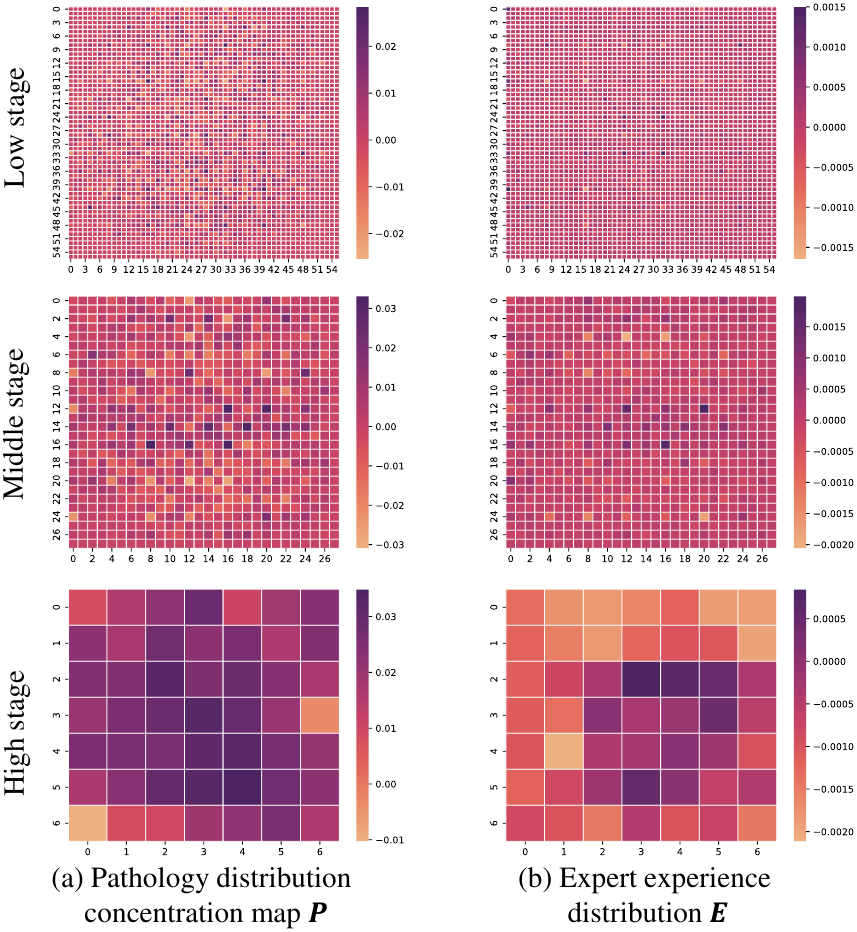}}
	\caption{Pathology distribution concentration map ($P$) weight distribution (first column) and expert experience 
distribution ($E$) (second column) at three stages of PCRNet18 for NC classification: low, middle and high.}
	\label{fig:6}
\end{figure}

\subsubsection{Visualizations of the Pathology Distribution 
Concentration Map $P$ and Expert Experience 
Distribution $E$}

Fig.~\ref{fig:6} plots pathology distribution concentration map ($P$) and expert experience distribution ($E$) of PCRNet18 at three stages for NC recognition on the CASIA2 NC dataset. We observe that PCRNet is inclined to set larger weights for the pathology attention of the up region than those of the down region, which becomes more evident at the high stage (On the top of Fig.~\ref{fig:6}. We see that expert experience distribution (On the bottom of Fig.~\ref{fig:6}) is complementary to pathology distribution concentration map, which can make a network locate informative regions precisely. The visual results also demonstrate that PCRNet keeps consistent with the clinical diagnosis mode by transforming clinical priors into learnable weights and biases.

\begin{figure}  
	\centering	
 \centerline{\includegraphics[width=1.0\linewidth]{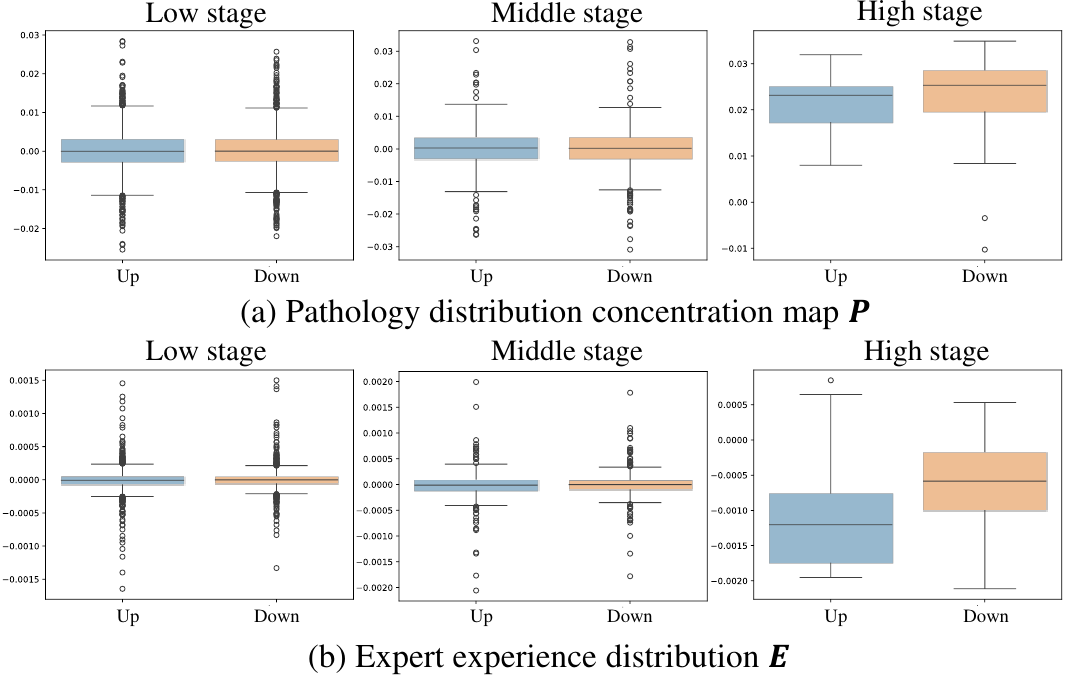}}
	\caption{Pathology distribution concentration map ($P$) and expert experience distribution ($E$) statistics of up- (first row) and bottom- (second row) feature map regions based on the PCRNet18 at three stages of a network for NC classification on CASIA2 NC dataset: low, middle, and high.}	
	\label{fig:7}
\end{figure}

Fig.~\ref{fig:7} offers pathology distribution concentration map $P$ statistics distribution and expert experience distribution $E$ statistics distribution of up- and down- feature map regions. Interestingly, we find that their distributions are similar, allowing a network to emphasize the down regions for learning informative feature representations as we expect \cite{doi:10.1076/opep.9.2.83.1523}, keep consistent with clinical research.

\begin{figure}	 
	\centering	
 \centerline{\includegraphics[width=0.8\linewidth]{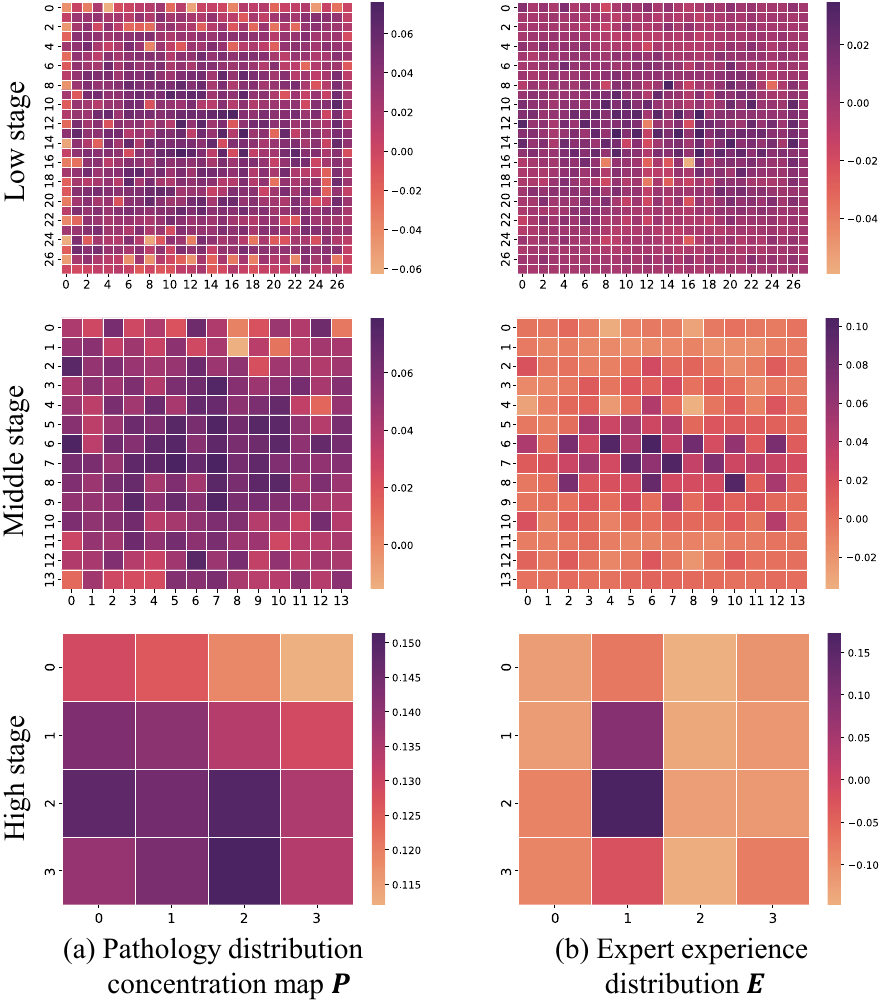}}
	\caption{Pathology distribution concentration map ($P$) distribution and expert experience ($E$) distribution of PCRNet at three stages for retinal disease classification on OCTMNIST dataset: low, middle and high. The first column denotes pathology context feature weight distribution, and the second column denotes expert experience distribution.}
	
	\label{fig:8}
\end{figure}

Fig.~\ref{fig:8} presents pathology distribution concentration map and expert experience distribution of PCRNet18 at three stages for retinal disease recognition on the OCTMNIST dataset. 
We also see that PCRNet18 is biased to set larger Pathology distribution concentration map and expert experience values for the central region, which is the pathological region of retinal diseases \cite{kermany2018identifying}. Specifically, expert experience values in the center feature map regions are larger than in other regions, proving that introducing expert experience as the diagnosis experience is complementary to pixel-wise context feature weight to guide a network to locate significant regions accurately.

\hspace{-2em}

\subsubsection{Visualizations of Pixel-wise Attention Map in PCRNet}
\begin{figure}  
    \centering	
 \centerline{\includegraphics[width=0.98\linewidth]{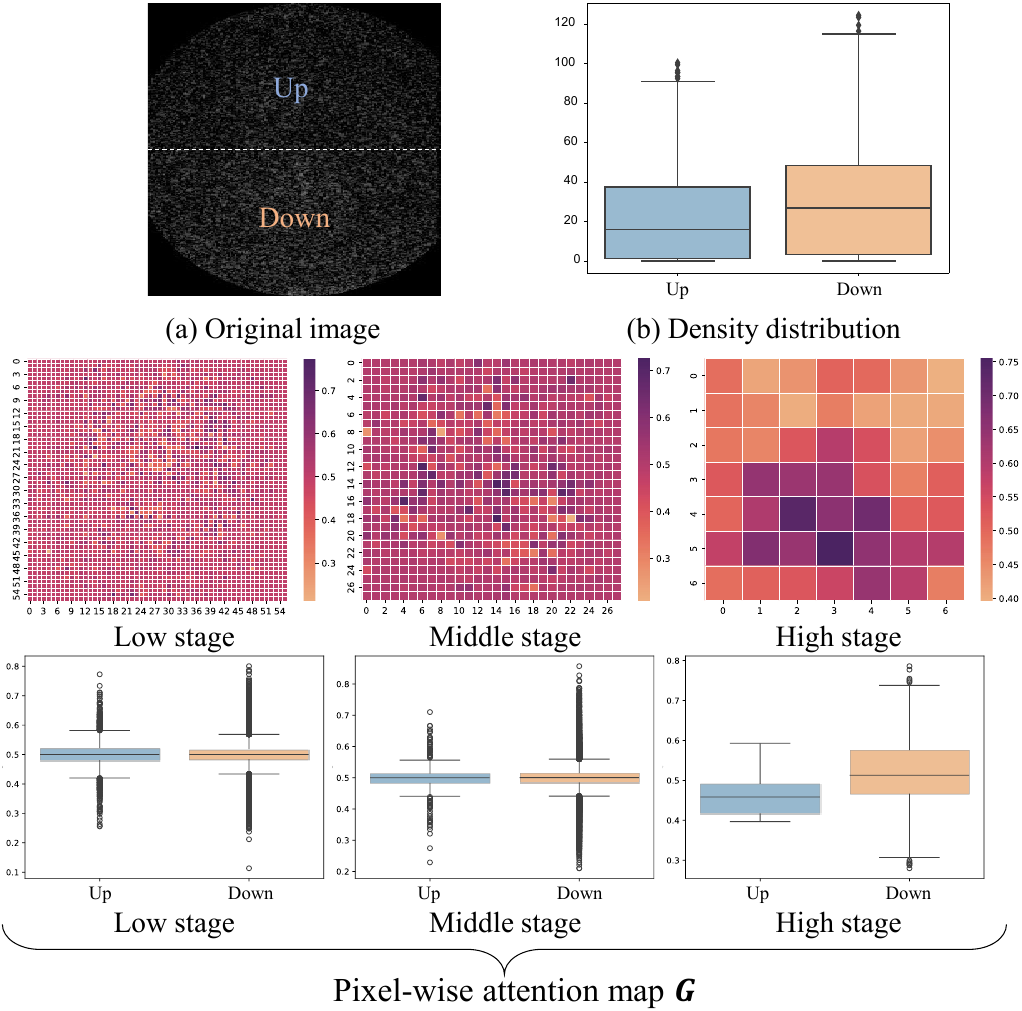}}
        \caption{The pixel-wise attention map ($G$) generated by PCRNet18 for NC classification task on CASIA2 NC dataset.
        First row: The original image with moderate cataract (left); the image density distribution of up- and down- regions.
	Second row: generated fine-turned feature map by PCRNet18 at three stages.
	Third row: pixel-wise attention map weight distribution of up- and down- regions.}
    \label{att_G}
\end{figure}

Fig.~\ref{att_G} presents a representational pixel-wise attention map ($G$) generated by PCRNet18 at three stages for an AS-OCT image of moderate NC. The feature map weight statistics distributions of down- and up- feature map regions are plotted to explain how PCRNet guides a network to focus on significant regions.
When the network goes deep, PCRNet18 prefers to generate large weights for down feature map regions. Specifically, the weights distribution gap between down- and up- regions becomes more evident at the high stage, which is consistent with the density value distribution gap between up and down nucleus regions, as shown in the first row of Fig.~\ref{att_G}. The attention weight distribution of PCRNet18 agrees with the clinical diagnosis mode that a clinician intends to make diagnosis results by pathology context and his diagnosis experience \cite{doi:10.1076/opep.9.2.83.1523}, demonstrating our PCRNet takes advantage of the clinical priors efficiently for improving interpretability of deep networks.

\hspace{-2cm}

\subsubsection{Visualization of Confusion Matrix on CASIA2 NC Dataset}

\begin{figure*}
	\centering	
 \centerline{\includegraphics[width=0.96\linewidth]{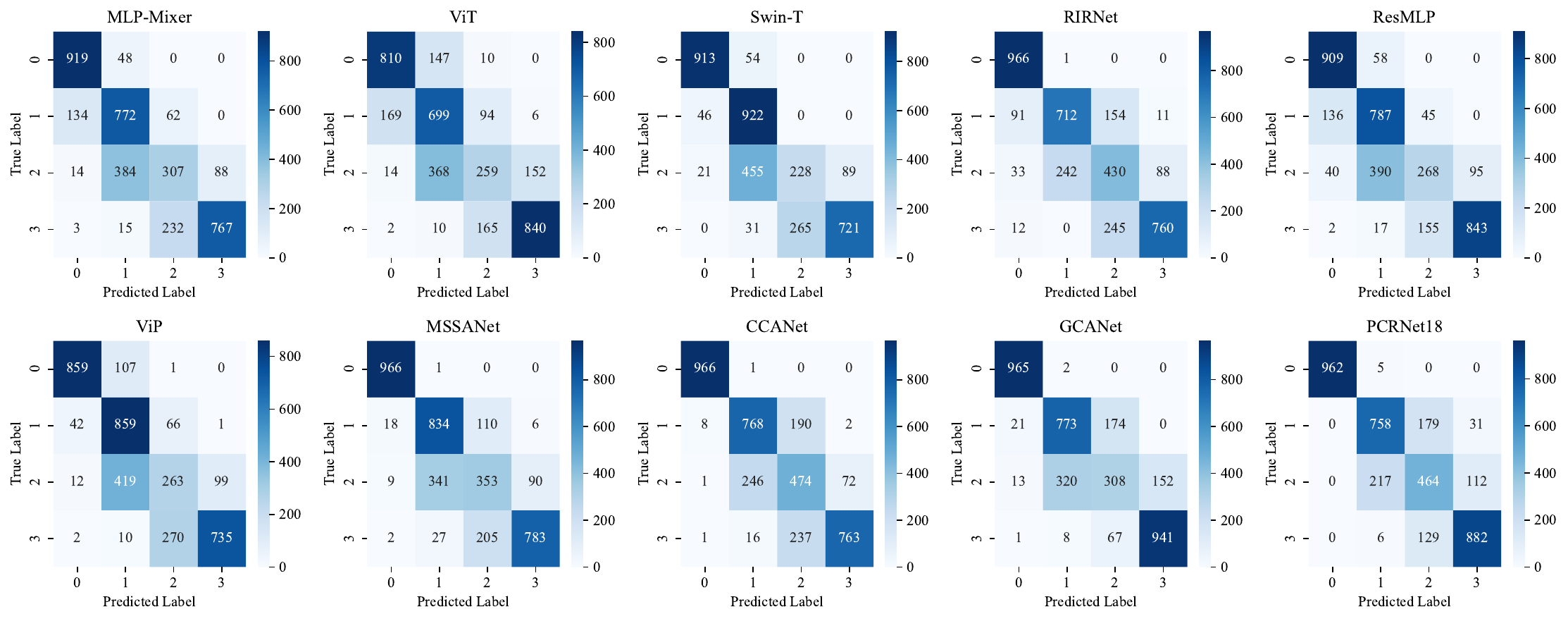}}
	\caption{The confusion matrix visualization for state-of-the-art deep neural networks (including MLP-Mixer, ViT, Swin-T, RIRNet, ResMLP, ViP, MSSANet, CCANet, and GCANet) and our proposed PCRNet18 on the CASIA2 NC dataset.}
\label{cm}
\end{figure*}

Figure~\ref{cm} presents the confusion matrix of PCRNet18 and SOTA DNNs on the CASIA2 NC dataset, providing more details on their performance in each severity level. Interestingly, all methods achieve relatively good performance in the normal (label 0) and severe (label 3) classes, while most misclassifications occur between the mild (label 1) and moderate (label 2) classes. Notably, accurately diagnosing these intermediate stages is crucial for surgical planning. 
Our proposed PCRNet integrates expert priors, effectively enhancing its discriminative ability between mild and moderate cases. This improvement further demonstrates the advantages of incorporating expert knowledge into deep learning models for more reliable diagnosis.

\subsubsection{Visualization of Grad-CAM Images on CASIA2 NC Dataset}

\begin{figure*}
	\centering	
 \centerline{\includegraphics[width=1.0\linewidth]{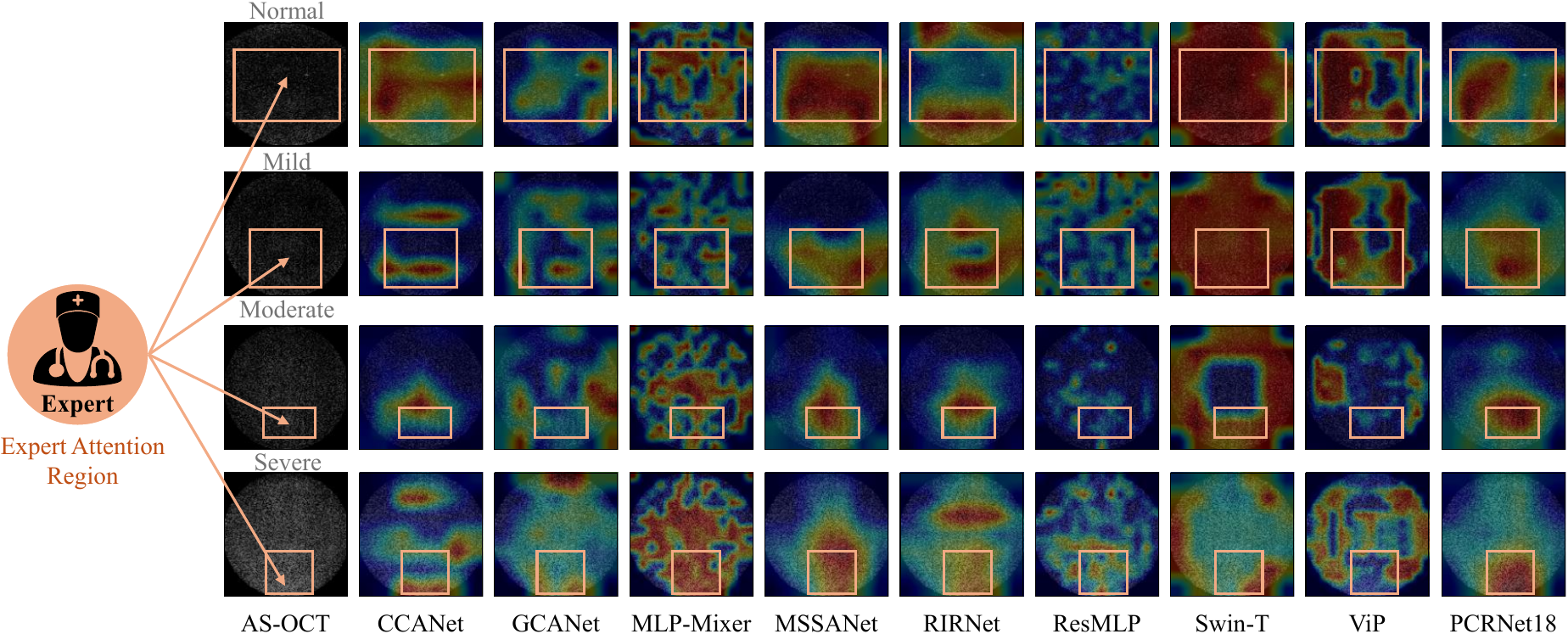}}
	\caption{Comparison between the Grad-CAM visualizations of various state-of-the-art deep neural networks (CCANet, GCANet, MLP-Mixer, MSSANet, RIRNet, ResMLP, Swin-T, and ViP) and the proposed PCRNet18 on the CASIA2 NC dataset, where the expert attention regions are labled.}
\label{gradcam}
\end{figure*}

We visualize the Grad-CAM images generated by the proposed PCRNet18 and advanced SOTA DNNs on the CASIA2 NC dataset to highlight the regions where the models focus. Figure~\ref{gradcam} presents representative examples across different NC severity levels.
PCRNet effectively localizes opacity by integrating expert priors, focusing on the entire nuclear region for normal cases and shifting toward the lower nucleus as severity increases, and exhibits attention patterns more consistent with the expert attention regions.
In contrast, ViP, Swin-T, and MLP-Mixer fail to capture these key regions, resulting in lower performance, consistent with the results in Table~\ref{NC}. 
These results underscore the advantage of expert priors in enhancing PCRNet’s diagnostic performance.

\subsubsection{Visualization of Pathology Attention under Different Coefficients $\mu$}

\begin{figure*}
	\centering	
 \centerline{\includegraphics[width=0.8\linewidth]{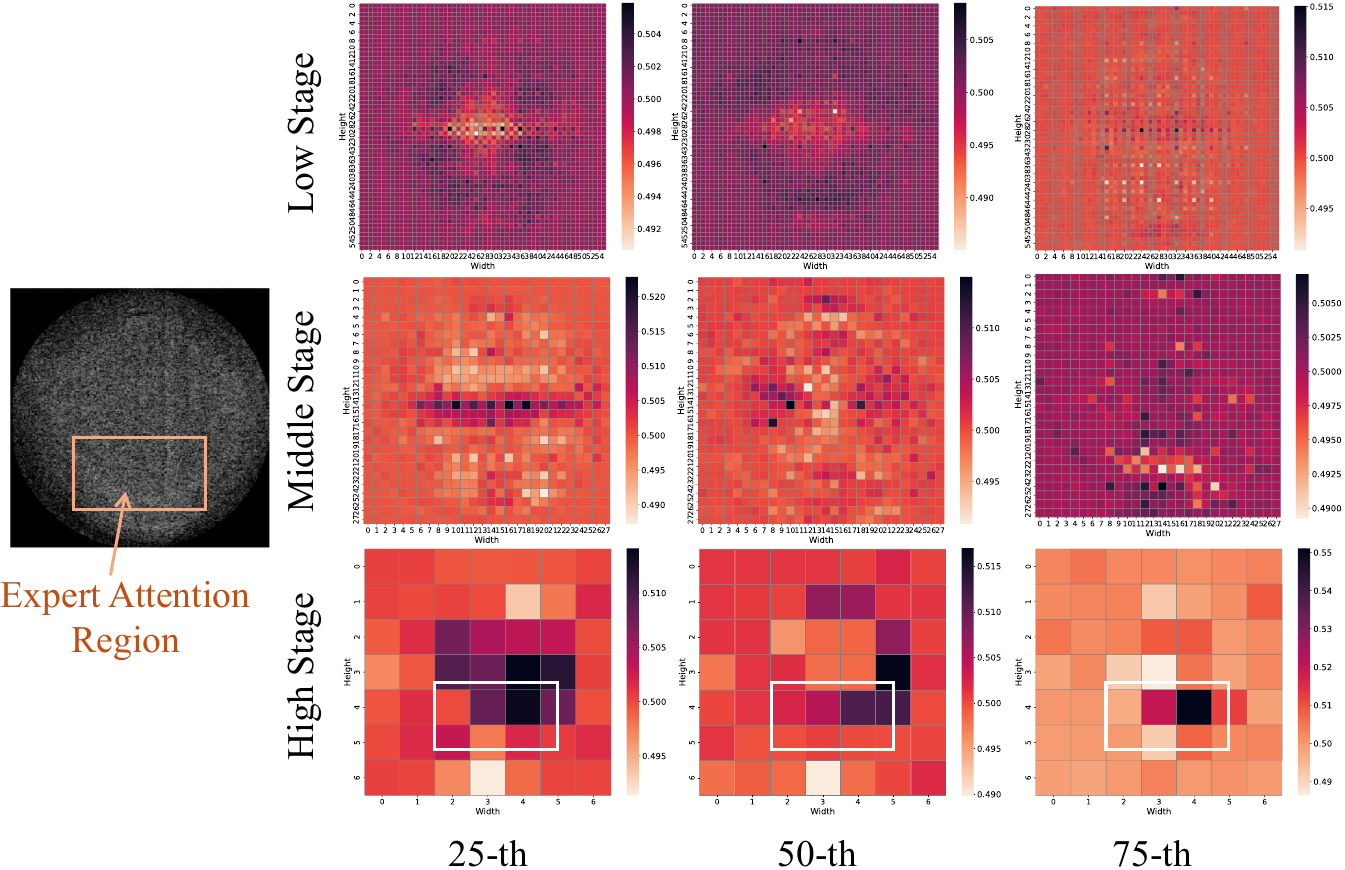}}
	\caption{Visualization of pathology attention heatmaps of the proposed PCRNet18 with different coefficient values $\mu$ (25-th, 50-th, and 75-th) on the CASIA2 NC dataset.}
\label{qss_att}
\end{figure*}

We visualize the attention maps $G$ of the proposed PCRNet18 on the CASIA2 NC dataset under different $\mu$ values to examine the effect of the Quantile Statistics Sampling (QSS) method.
Figure~\ref{qss_att} presents representative examples illustrating the relationship between the model’s attention regions and the selected percentiles. In the high-stage attention maps, the model’s attention becomes more focused and consistent with the expert at the 75th percentile, consistent with the results in Table~\ref{respond}.

\section{Limitations and Future Work}

This paper argues that pathology context and expert experience priors play vital roles in ocular disease diagnosis, but previous works ignore exploring these clinical priors to improve ocular disease recognition performance and decision-making interpretability. To address this gap, we propose two external prior infusion blocks, PRM and EPGA, to incorporate pathology context and expert experience priors into DNN representations. By combining PRM and EPGA into DNN, we construct the PCRNet for automated ocular disease recognition. Additionally, we propose an Integrated Loss (IL) function that accounts for sample-wise loss distributions and label frequency imbalance to further enhance performance. While our method outperforms recent SOTA methods across three ocular disease diagnosis tasks, it still has certain limitations, as discussed below:

\begin{itemize}

\item PCRNet primarily incorporates external clinical priors through the pathology distribution concentration map ($P$) and expert experience distribution ($E$) in a pixel-wise manner. However, such a formulation may be insufficient for capturing more complex clinical prior information, such as structural or functional characteristics.

\item We propose the Integrated Loss (IL) to enhance the ocular disease recognition performance of PCRNet, which still has room for improvement. For example, IL could be combined with other self-supervised methods to further enhance the representational capacity of DNNs using large amounts of unlabeled data.

\item We only evaluate the effectiveness of PCRNet on ocular disease classification tasks and single-modality settings due to dataset limitations.
\end{itemize}
To address the above limitations, we plan to enhance the architectural design of PCRNet by introducing expert Mixture-of-Experts (MoE) approaches, aiming to better capture diverse and complex clinical priors. Furthermore, we will expand our experiments to include additional clinical diagnostic tasks and collect multi-modality datasets to more comprehensively evaluate the generalization ability and effectiveness of our method.

\section{Conclusion}
\label{sec:conclude}
In this paper, we construct a PCRNet to recognize ocular diseases based on ophthalmic images. In PCRNet, this paper proposes Pathology Recalibration Module (PRM) and Expert Prior Guidance Adapter (EPGA), which is the first time to explicitly leverage the potential of the clinical priors of pathology context and expert experience for improving ocular disease recognition performance and explanation of DNNs from two different clinical prior-driven perspectives. Moreover, we also introduce an integrated loss (IL) to further boost the performance of our PCRNet. The extensive experiments on three ocular disease datasets demonstrate that our PCRNet and IL achieve better ocular disease recognition results than SOTA methods. In addition, we verify the capability of PCRNet in focusing on significant pixel-wise context locations and suppressing useless ones by visualizing pixel-wise context feature distributions, bias weights, and attention weights, which conduces to a better understanding of PCRNet and agrees with the clinical diagnosis mode. We hope our work may inspire the research of exploiting clinical priors to improve the interpretability of DNNs, prompting clinical applications of DNN-based computer aid diagnosis techniques.

\backmatter

\section*{Acknowledgements}
This work was supported in part by National Key R\&D Program of China (No.2024YFC2510800) and Shenzhen Medical Research Funds (Grant No. D2402014).

\section*{Declaration of Competing Interest}
The authors declare that they have no known competing financial interests or personal relationships that could have appeared to influence the work reported in this paper.

\noindent

\bibliography{cas-refs}

\end{document}